\documentclass[10pt,twocolumn,letterpaper]{article}

\usepackage{cvpr}              

\usepackage{multirow}
\usepackage{booktabs}
\usepackage{makecell}
\usepackage{pifont}
\usepackage{xcolor}
\usepackage{color}
\definecolor{cvprblue}{rgb}{0.21,0.49,0.74}
\usepackage[pagebackref,breaklinks,colorlinks,allcolors=cvprblue]{hyperref}
\usepackage{amsmath}
\usepackage{amsfonts}
\usepackage{tikz}
\usepackage{graphicx}
\usepackage{amssymb}
\usepackage{balance}
\usepackage{url}            
\usepackage{bigstrut}
\usepackage{times}
\usepackage{microtype} %
\usepackage{wrapfig}
\usepackage{balance} 
\usepackage{algorithm}
\usepackage{algorithmic}
\usepackage{stackengine}
\usepackage{threeparttable}
\usepackage{array}

\title{$\text{S}^{3}$Mamba: Arbitrary-Scale Super-Resolution via Scaleable State Space Model}

\author{Peizhe Xia$^{1,2}$\textsuperscript{\textdagger}  ~~~Long Peng$^{1,2}$\textsuperscript{\textdagger}  ~~~Xin Di$^{1,2}$ ~~~Renjing Pei$^{2*}$ ~~~Yang Wang$^{1}$\thanks{Corresponding Authors: Renjing Pei, \texttt{peirenjing@huawei.com}; Yang Wang, \texttt{ywang120@ustc.edu.cn}. \textsuperscript{\textdagger} These authors contributed equally to this work. This research was done while Peizhe Xia and Long Peng were interns at Huawei Noah’s Ark Lab.} \\~~~Yang Cao$^{1}$ ~~~Zheng-Jun Zha$^{1}$ \\%
{$^{1}$University of Science and Technology of China ~~~$^{2}$Huawei Noah’s Ark Lab}\\
{\small \textbf{\url{https://github.com/xiapeizhe12138/S3Mamba-ArbSR}}}
}
\begin{document}

\maketitle
\begin{abstract}
{Arbitrary scale super-resolution (ASSR) aims to super-resolve low-resolution images to high-resolution images at any scale using a single model, addressing the limitations of traditional super-resolution methods that are restricted to fixed-scale factors (e.g., \( \times2 \), \( \times4 \)). The advent of Implicit Neural Representations (INR) has brought forth a plethora of novel methodologies for ASSR, which facilitate the reconstruction of original continuous signals by modeling a continuous representation space for coordinates and pixel values, thereby enabling arbitrary-scale super-resolution. Consequently, the primary objective of ASSR is to construct a continuous representation space derived from low-resolution inputs. However, existing methods, primarily based on CNNs and Transformers, face significant challenges such as high computational complexity and inadequate modeling of long-range dependencies, which hinder their effectiveness in real-world applications. To overcome these limitations, we propose a novel arbitrary-scale super-resolution method, called $\text{S}^{3}$Mamba, to construct a scalable continuous representation space. Specifically, we propose a Scalable State Space Model (SSSM) to modulate the state transition matrix and the sampling matrix of step size during the discretization process, achieving scalable and continuous representation modeling with linear computational complexity. Additionally, we propose a novel scale-aware self-attention mechanism to further enhance the network's ability to perceive global important features at different scales, thereby building the $\text{S}^{3}$Mamba to achieve superior arbitrary-scale super-resolution. Extensive experiments on both synthetic and real-world benchmarks demonstrate that our method achieves state-of-the-art performance and superior generalization capabilities at arbitrary super-resolution scales. The code will be publicly available.
}

\end{abstract}    

\section{Introdution}

{With the rapid advancement of digital imaging technology and computational photography~\cite{peng2020cumulative,wang2023decoupling,peng2021ensemble,peng2024lightweight,wang2023brightness,yi2021structure,yi2021efficient,wu2024rethinking}, image super-resolution (SR) has become a significant research topic in computer vision and image processing~\cite{gunturk2004super,zou2011very,shi2013cardiac,ren2024ninth,conde2024real,li2023ntire,peng2024towards}. SR aims to reconstruct high-resolution (HR) images from low-resolution (LR) inputs to enhance visual quality. However, traditional factor-fixed SR methods~\cite{lim2017edsr,zhang2018rcan,liang2021swinir,peng2024efficient,chen2022hat,liu2018NLRN,mei2021NLSA} often can only upscale LR images by fixed magnification factors\cite{shi2016pixelshuffle,zhang2020RDNIR,dai2019SAN,niu2020HAN}, such as ($\times$2, $\times$3, $\times$4), which makes it difficult to meet the demands of real-world applications that require arbitrary magnification. Consequently, arbitrary scale super-resolution (ASSR) has been proposed and has garnered widespread attention, effectively achieving any super-resolution scale using a single model.}
\begin{figure*}[t]
 \vspace{-5mm}
  \centering
  \includegraphics[width=1\linewidth]{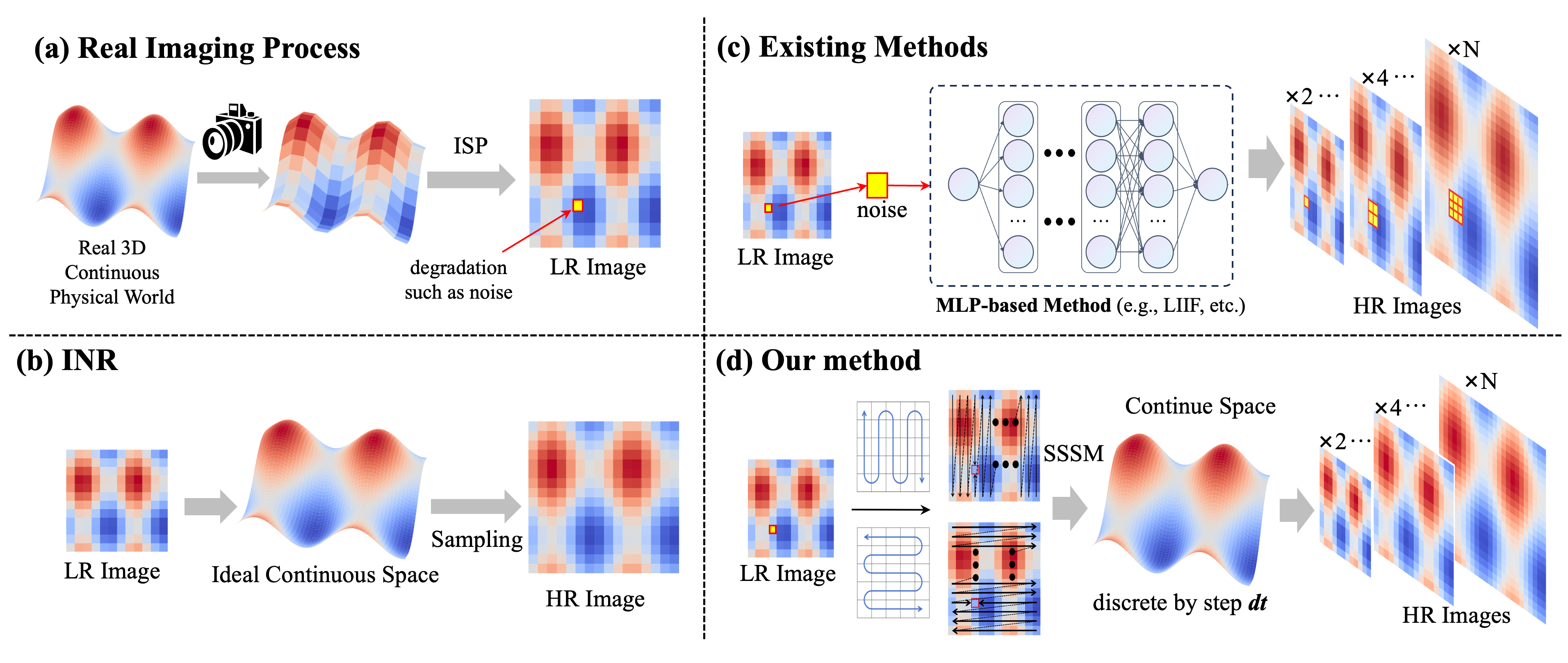}
   \vspace{-8mm}
\caption{(a) During real-world imaging, the continuous 3D physical world is discretized into an image through cameras and ISPs, resulting in an LR image due to sensor resolution. (c) Existing MLP-based INR methods often use point-to-point learning, making them susceptible to degradation such as noise. Additionally, the limited receptive field of MLPs cannot construct a perfect continuous space, as shown in (b). In contrast, our method (d) leverages scalable SSM to better capture global historical information and, through scalable training, reconstructs a continuous space more effectively, achieving superior arbitrary-scale super-resolution.}
 \vspace{-3mm}
  \label{fig:motivation}
\end{figure*}

{In reality, our physical world is three-dimensional and continuous. To record the physical world, various imaging devices have been invented to discretize signals by capturing reflected photons from the real world to obtain observable digital images, as shown in Figure~\ref{fig:motivation} (a). The limited quality and resolution of sensors result in low-quality low-resolution images~\cite{chen2024towards,wu2024mixnet,lim2017edsr,chen2021liif}. Therefore, the biggest challenge of ASSR is to learn the continuous signals of the real world from these discretized low-resolution images~\cite{chen2021liif,LTE,cao2023ciaosr}. Numerous approaches have been proposed~\cite{hu2019metasr,chen2021liif,lee2022lte,xu2021ultrasr,cao2023ciaosr,wei2023super,chen2023cascaded,LTE} to achieve this. Among these, implicit neural representation (INR) stands out as the most prominent and effective. INR constructed a mapping from continuous pixel coordinates and the discredited low-resolution images to the continuous high-resolution signal, achieving scalable super-resolution, as shown in Figure~\ref{fig:motivation} (b). }

{Numerous INR-based arbitrary-scale super-resolution methods have been proposed, achieving significant progress~\cite{hu2019metasr,chen2021liif,lee2022lte,xu2021ultrasr,cao2023ciaosr,wei2023super,chen2023cascaded,LTE}. For instance, LIIF~\cite{chen2021liif} is the first to introduce INR into arbitrary-scale super-resolution, utilizing multi-layer perceptrons (MLP) to reconstruct continuous mappings for arbitrary-scale enlargement. This approach has achieved impressive visual results and garnered significant attention. Following this, LTE~\cite{lee2022lte} and LINF~\cite{yao2023local} attempt to enhance performance by incorporating frequency domain information in the decoder. However, the limited receptive field and point-to-point learning approach of MLP make it difficult to capture contextual information, leading to challenges in constructing continuous HR images and making them susceptible to noise interference. This constrains the performance of INR in ASSR, as shown in Figure~\ref{fig:motivation} (c). Therefore, CiaoSR~\cite{cao2023ciaosr} and CLIT~\cite{chen2023cascaded} utilized Transformers to model global information, significantly improving model performance. However, although Transformers excel at modeling relationships among all tokens to capture contextual information, their self-attention mechanism incurs quadratic computational complexity. This quadratic increase in complexity with respect to input size makes them inefficient for real-world deployment. Therefore, there is an urgent need for an ASSR network capable of global modeling while maintaining high efficiency.}

To tackle the aforementioned challenges, we propose an innovative arbitrary-scale super-resolution method called $\text{S}^{3}$Mamba, which constructs a scalable continuous representation space, as shown in Figure~\ref{fig:motivation} (d). This approach introduces the State Space Model (SSM) into arbitrary-scale super-resolution for the first time. We further propose a novel Scalable State Space Model (SSSM) to modulate the state transition matrix and sampling step size during discretization, thus achieving scalable and continuous representation modeling with linear computational complexity. Additionally, we develop an advanced scale-aware self-attention mechanism to enhance the network's ability to capture globally significant features across various scales. These innovations culminate in the $\text{S}^{3}$Mamba, a versatile module that seamlessly integrates into diverse super-resolution backbones, thereby enhancing their efficacy at arbitrary scales. Comprehensive experiments on both real-world and popular synthetic benchmarks demonstrate our method's state-of-the-art performance, with superior generalization and continuous space representation capabilities in real-world scenarios. Our main contributions are as follows:

\begin{itemize}[leftmargin=3mm]

\item {We pioneer the introduction of the State Space Model into arbitrary-scale super-resolution and propose the novel Scalable State Space Model (SSSM). This model effectively modulates the state transition matrix and sampling step size during discretization, achieving scalable and continuous representation modeling with linear computational complexity.}
    
\item {We develop the $\text{S}^{3}$Mamba, introducing an innovative scale-aware self-attention mechanism that incorporates the SSSM. This enhancement significantly boosts the network's ability to capture globally significant features across various scales, ensuring superior performance at arbitrary scales.}

\item {Extensive experiments demonstrate that our method achieves the best performance on the popular DIV2K benchmark and exhibits the best performance and generalization capabilities on real-world COZ benchmarks.}
    
\end{itemize}

\section{Related work}
\subsection{Arbitrary-Scale Super-Resolution}

{Different from traditional fixed-scale single image super-resolution~\cite{dong2014srcnn,ledig2017srresnet,zhang2018rcan,kim2016vdsr,cavigelli2017cas,zhang2021DPIR,wang2018esrgan}, Arbitrary-Scale Super-Resolution (ASSR) has the ability to enhance image quality and resolution across various scales, garnering significant attention in the fields of image processing and computer vision~\cite{cao2023ciaosr,hu2019metasr,fu2024continuous}. For example, MetaSR first proposed a meta-upscale module to tackle this challenge~\cite{hu2019metasr}. Inspired by the success of implicit neural representation (INR) in 3D shape reconstruction\cite{sitzmann2020SIREN,chen2019learning,michalkiewicz2019implicit,gropp2020implicit,sitzmann2019scene,mildenhall2021nerf}, the LIIF method employs a multilayer perceptron (MLP) to learn a continuous representation of the image~\cite{chen2021liif}. It takes continuous image coordinates and surrounding image features as input, outputting the RGB values at given coordinates. However, MLP has limitations in learning high-frequency components. LTE~\cite{lee2022lte} addresses this issue by effectively encoding image textures in the Fourier space. SRNO~\cite{wei2023super} introduces neural operators to capture global relationships within the image. ITSRN~\cite{yang202itsrn} further innovatively proposes an implicit transformer based on INR structures to fully leverage screen image content. Cao \textit{et al.} proposed CiaoSR as a continuous implicit attention network, that learns and integrates the weights of local features nearby, achieving the current state-of-the-art (SOTA) performance~\cite{cao2023ciaosr}. These methods provide diverse pathways and possibilities for achieving arbitrary-scale super-resolution. LMF~\cite{he2024latent} optimized image representation by reducing MLP dimensions and controls rendering intensity through modulation to reduce the computational cost of the upsampling module. COZ~\cite{fu2024continuous} provided a benchmark for real-world scenarios, offering a dataset for arbitrary-scale super-resolution tasks captured in real scenes, along with a lightweight INR network. However, the aforementioned ASSR methods primarily utilize MLPs for point-to-point generation of high-resolution image pixels, which tends to overlook the intrinsic continuity within images. This oversight makes them susceptible to degradation artifacts, resulting in unrich detail and artifacts.} 

\subsection{State Space Models}
{State Space Models (SSMs) were first developed in the 1960s for control systems~\cite{a:25}, providing a framework for modeling systems with continuous signal inputs. In recent times, the evolution of SSMs has facilitated their integration into the realm of computer vision~\cite{a:30,patro2024simba,fu2024ssumamba,chen2024changemamba}. A prominent example is Visual Mamba, which introduced a residual VSS module and implemented four scanning directions. This innovation resulted in superior performance over ViT~\cite{a:31}, while maintaining a lower model complexity, thus garnering considerable attention~\cite{c:8,a:32,a:33,a:34,a:35,a:36,yan2024mambasr,xiao2024frequency,di2024qmambabsr}. Notably, MambaIR~\cite{c:8} pioneers the use of SSMs in image restoration, boosting both efficiency and global perceptual capability. Despite these advances, the potential of the continuous representation modeling ability of SSM in arbitrary-scale super-resolution tasks remains underexplored. Therefore, we propose a novel scalable State Space State, which leverages the continuous state space of SSMs to enhance the network's capability in continuous representation, thereby achieving high-quality continuous arbitrary-scale super-resolution.}

\begin{figure*}[ht]
 \vspace{-3mm}
  \centering
  \includegraphics[width=\linewidth]{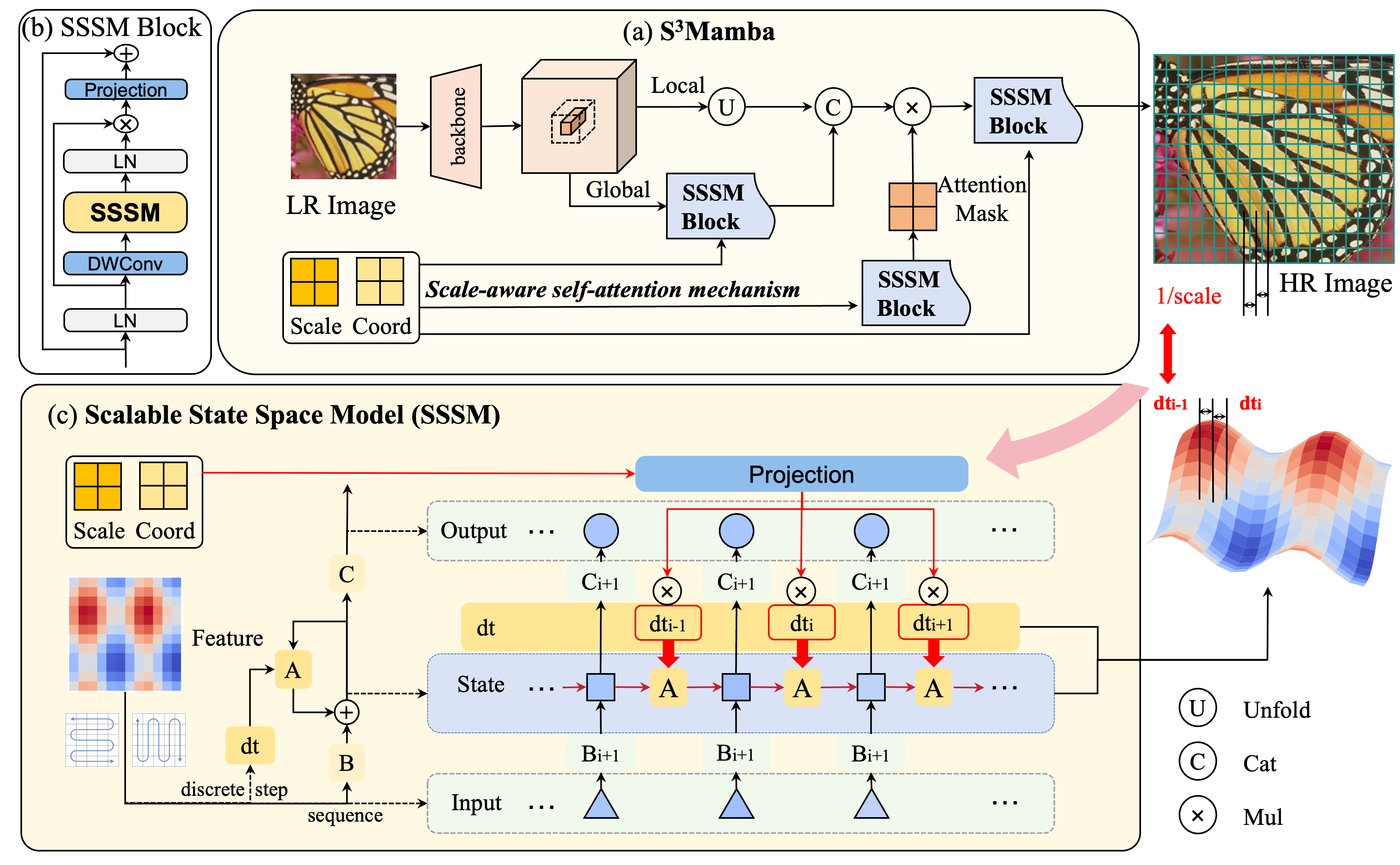}
   \vspace{-3mm}
\caption{(a) Illustration of the proposed $\text{S}^3$Mamba framework. (b) The SSSM Block consists of the SSSM, along with multiple instance normalization layers, depthwise convolution (DWConv), and projection layers. (c) The Scalable State Space Model (SSSM) is proposed to modulate the state transition matrix and the sampling matrix of step size during the discretization process, achieving scalable and continuous representation modeling with linear computational complexity.}
 \vspace{-3mm}
  \label{fig:Method}
\end{figure*}
\section{Preliminary and Motivation}
Our three-dimensional, continuous physical world is recorded by cameras that convert reflected photons into digital images~\cite{son2012three}, as shown in Figure~\ref{fig:motivation} (a). However, limitations in CMOS and CCD sensor technology result in low-resolution images, failing to meet consumers' demands for higher resolution and better quality~\cite{wang2020deep,chen2022real}. Image super-resolution techniques have been developed to generate high-resolution (HR) images from low-resolution (LR) counterparts. Unlike traditional fixed-scale methods, Arbitrary-Scale Super-Resolution (ASSR) aims to reconstruct the original continuous scene, generating HR images at any resolution. The main challenge of ASSR is learning continuous signals from discretized data~\cite{liu2024arbitrary,chen2021learning}, as shown in Figure~\ref{fig:motivation} (b). The implicit neural representation (INR) stands out as the most prominent and effective in ASSR. By learning the mapping from pixel coordinates to pixel values, INR is able to generate scalable, high-quality high-resolution images, as formulated:
\begin{equation}
F_{LR} = \Psi(LR) 
\end{equation}
\begin{equation}
HR^{RGB}_{(i,j)} = \phi(F_{{LR}}, {coord}_{(i,j)}, {scale})
\end{equation}
where, $F_{LR}$ represents the features of the low-resolution image $LR$, and $\Psi$ denotes the feature extractor. ${coord}$ represents the coordinates location, ${scale}$ indicates the magnification factor, and $HR^{RGB}$ denotes the high-resolution RGB image. The goal of Implicit Neural Representation (INR) is to learn a continuous function \( \phi \) that maps coordinates and images to continuous signals, effectively mapping different scales of the same scene into a unified continuous representation space. Various INR methods, like LIIF~\cite{chen2021liif} and LTE~\cite{lee2022lte}, use a Multi-Layer Perceptron (MLP) for ASSR, but MLPs' limited receptive field and point-to-point approach ignore contextual and historical data, leading to vulnerability to noise and poor continuous representation capability, as shown in Figure~\ref{fig:motivation} (c). An intuitive approach is to introduce Transformers to capture global information, as seen in methods like CiaoSR~\cite{cao2023ciaosr}. While Transformers effectively capture context, their quadratic computational complexity makes them impractical for real-world applications. More detailed analyses and comparisons are provided in the supplementary material.

\section{Method}

To reconstruct a scalable continuous representation space, we propose a novel arbitrary-scale super-resolution method called $\text{S}^{3}$Mamba, as shown in Figure~\ref{fig:Method} (a). This approach leverages the Scalable State Space Model (SSSM) to adaptively capture global and scale-dependent features, ensuring consistent continuous representations across varying scales. Additionally, our innovative scale-aware self-attention mechanism is introduced to enhance the network's ability to perceive globally significant features at different scales, thereby reconstructing high-quality high-resolution images efficiently and effectively.

\subsection{Proposed Scalable State Space Model}
To better capture global historical information without incurring significant computational overhead, we turn our attention to state space models. Benefiting from the linear complexity and global modeling capacity of state space models, we introduce the State Space Model into arbitrary-scale super-resolution for the first time. Let's briefly review the State Space Model (SSM). The latest advances in structured state space models (S4) are largely inspired by continuous linear time-invariant (LTI) systems, which map input \( x(t) \) to output \( y(t) \) through an implicit latent state \( h(t) \in \mathbb{R}^N \) ~\cite{c:8}. This system can be represented as a linear ordinary differential equation (ODE):
\begin{equation}
\begin{aligned}
    \dot{h}(t) &= Ah(t) + Bx(t), \\
    y(t) &= Ch(t) + Dx(t).
\end{aligned}
\label{eq:1}
\end{equation}
where \(N\) is the state size, \(A \in \mathbb{R}^{N \times N}\), \(B \in \mathbb{R}^{N \times 1}\), \(C \in \mathbb{R}^{1 \times N}\), and \(D \in \mathbb{R}\). To adapt to digital information processing, the continuous function of the state space model in Eq.~\ref{eq:1} is discretized into a sequence analysis model. Specifically, the state space model uses a zero-order hold as follows: 
\begin{equation}
\begin{aligned}
    \overline{A} &= \exp(\Delta A), \\
    \overline{B} &= (\Delta A)^{-1} (\exp(\Delta A) - I) \Delta B,
\end{aligned}
\label{eq:2}
\end{equation}
In this process, the sampling interval $\Delta$ determines the arrangement of discrete signals, so within the SSM, $\Delta$ dictates the correlation and association between adjacent inputs. Finally, we arrive at the discrete state space representation, as shown in the following equations:
\begin{equation}
\begin{aligned}
    h_k &= \overline{A} h_{k-1} + \overline{B} x_k, \\
    y_k &= Ch_k + Dx_k,
\end{aligned}
\label{eq:3}
\end{equation}

In the traditional state space model, \(\Delta\) is determined solely by the current input, making it well-suited for scale-invariant vision tasks. However, because the actual physical distance between adjacent pixels varies with different scales of the same scene, the correlation and association between adjacent pixels also change with scale. An INR model trained solely on a traditional SSM may fail to capture these scale-dependent patterns, resulting in different continuous representations for different scales of the same scene. This is inconsistent with the fundamental goal of INR. 

To address this issue, we propose a novel Scalable State Space Model (SSSM), which incorporates scale and continuous coordinate information into the state space equations of the state space model and adjusts \(\Delta_{x_k}\) to achieve scale awareness. Specifically, we use a learnable MLP layer to input the scale, generating a scale modulation factor for each time step, which is then introduced into the current \(\Delta_{x_k}\), as formulated:
\begin{align}
\Delta_{x_k} &= \omega(x_k) ,\\
\Delta^{scale}_{x_k} &= \sigma({scale}, {coord}_{x_k}), \\
\Delta^{'}_{x_k} &= \Delta_{x_k} \cdot \Delta^{scale}_{x_k}.
\label{eq:5}
\end{align}
where $\omega$ and $\sigma$ represent multilayer perceptron layers. This approach allows the SSSM to adaptively adjust the interaction patterns of adjacent points at different scales. This ensures consistency in network outputs for the same input across various scales, allowing our arbitrary-scale super-resolution model to maintain a consistent continuous representation space when handling data of different output sizes.

Furthermore, in the original state space equations, the parameter matrix \( B \) represents the mapping pattern from the input to the state space. It is directly determined by the current input to produce \( B_{x_k} \), which can still prevent the SSM-based upsampling module from effectively capturing continuous space representation methods at different scales. Therefore, we follow Eq.~\ref{eq:5} to transform the same process to matrix \( B_{x_k} \) into \( B^{'}_{x_k} \), allowing it to better perceive the mapping equations across different scales. This ensures that the state space model can adapt to any magnification level. The above process can be formulated as:
\begin{align}
B_{x_k}, C_{x_k}, \Delta_{x_k} &= \omega(x_k) ,\\
B^{scale}_{x_k}, \Delta^{scale}_{x_k} &= \sigma({scale}, {coord}_{x_k}), \\
\Delta^{'}_{x_k} &= \Delta_{x_k} \cdot \Delta^{scale}_{x_k}, \\
B^{'}_{x_k} &= B_{x_k} \cdot B^{scale}_{x_k}.
\label{eq:6}
\end{align}
Then, the discretization process of our Scalable State Space Model (SSSM) can be formulated as:
\begin{equation}
\begin{aligned}
    \overline{A}^{'}_{x_k} &= \exp(\Delta^{'}_{x_k} A), \\
    \overline{B}^{'}_{x_k} &= (\Delta^{'}_{x_k} A)^{-1} (\exp(\Delta^{'}_{x_k} A) - I) \Delta^{'}_{x_k} B^{'}_{x_k},
\end{aligned}
\label{eq:7}
\end{equation}
Finally, the discretized state space equations of our SSSM can be represented by the following equations:
\begin{equation}
\begin{aligned}
    h_k &= \overline{A}^{'}_{x_k} h_{k-1} + \overline{B}^{'}_{x_k} x_k, \\
    y_k &= C_{x_k}h_k + Dx_k.
\end{aligned}
\label{eq:8}
\end{equation}
Through the aforementioned design, we follow to~\cite{a:30} to construct the SSSM block to adeptly capture scale variations, as shown in Figure~\ref{fig:Method} (b) and (c). This allows low-resolution images, sampled at different scales within a unified continuous scene, to be represented within a single continuous space. This capability facilitates the construction of an enhanced continuous space, yielding high-resolution images across arbitrary scales that are visually pleasing and rich in detail.

\begin{table*}[ht]
\caption{Quantitative comparison with state-of-the-art methods for arbitrary-scale SR on the \underline{\textbf{real-world COZ  set}} (PSNR (dB) / SSIM). \textbf{Bold} indicates the best performance. Out-of-scale means the models were not trained with these large scales.
}
\label{tab:comp_COZ}
\centering
\resizebox{1\textwidth}{!}{
\begin{tabular}{l||l||ccc||ccc}
\toprule
& & \multicolumn{3}{c||}{$\qquad$In-scale$\qquad$} & \multicolumn{3}{c}{Out-of-scale} \\
\cmidrule{3-8}
\multirow{-2.5}{*}{Backbones} & \multirow{-2.5}{*}{Methods} &~~~~~$\times3$~~~~~ & ~~~~~$\times3.5$~~~~~ & ~~~~~$\times4$~~~~~ & ~~~$\times5$~~~ & ~~~$\times5.5$~~~ & ~~~$\times6$~~~ 
\\
\hline\hline
\multirow{1}{*}{}

\multirow{9}{*}{EDSR~\cite{lim2017edsr}} &

MetaSR~\cite{Meta-SR} & 26.65/0.767 & 25.80/0.752 & 25.22/0.740 & 24.39/0.720 & 24.09/0.711 & 23.31/0.678 \\
& LIIF~\cite{chen2021liif} & 26.61/0.767 & 25.76/0.752 & 25.16/0.741 & 24.32/0.721 & 24.01/0.711 & 23.23/0.679 \\
& LTE~\cite{lee2022lte} & 26.55/0.767 & 25.71/0.752 & 25.15/0.740 & 24.37/0.720 & 24.05/0.712 & 23.26/0.679 \\
& LINF~\cite{yao2023local} & 26.53/0.762 & 25.66/0.750 & 25.10/0.737 & 24.29/0.719 & 23.99/0.711 & 23.21/0.677 \\
& SRNO~\cite{wei2023super} & 26.59/0.766 & 25.70/0.752 & 25.15/0.741 & 24.31/0.722 & 24.05/0.712 & 23.25/0.680 \\
& LIT~\cite{chen2023cascaded} & 26.58/0.766 & 25.71/0.753 & 25.16/0.741 & 24.35/0.721 & 24.00/0.712 & 23.19/0.679 \\
& CiaoSR~\cite{cao2023ciaosr} & 26.56/0.770 & 25.65/0.755 & 25.13/0.746 & 24.31/0.725 & 23.96/0.721 & 23.23/0.709 \\
& LMI~\cite{fu2024continuous} & 26.66/0.768 & 25.78/0.752 & 25.22/0.741 & 24.39/0.722 & 24.08/0.713 & 23.29/0.680 \\
& \bf{Ours} & \bf{26.71/0.773} & \bf{25.84/0.755} & \bf{25.27/0.746} & \bf{24.39/0.726} & \bf{24.09/0.723} & \bf{23.34/0.709} \\

\midrule
\multirow{9}{*}{RDN~\cite{zhang2018rdn}} &

MetaSR~\cite{Meta-SR} & 26.65/0.767 & 25.80/0.752 & 25.22/0.740 & 24.39/0.720 & 24.09/0.711 & 23.31/0.678 \\
& LIIF~\cite{chen2021liif} & 26.69/0.766 & 25.83/0.752 & 25.23/0.740 & 24.39/0.718 & 24.13/0.711 & 23.28/0.679 \\
& LTE~\cite{lee2022lte}  & 26.64/0.767 & 25.74/0.752 & 25.17/0.740 & 24.40/0.719 & 24.10/0.709 & 23.28/0.676 \\
& LINF~\cite{yao2023local} & 26.60/0.762 & 25.73/0.750 & 25.15/0.737 & 24.32/0.719 & 24.03/0.711 & 23.28/0.677 \\
& SRNO~\cite{wei2023super} & 26.67/0.766 & 25.73/0.752 & 25.19/0.741 & 24.40/0.722 & 24.09/0.712 & 23.28/0.680 \\
& LIT~\cite{chen2023cascaded} & 26.66/0.766 & 25.79/0.753 & 25.19/0.741 & 24.36/0.721 & 24.03/0.712 & 23.25/0.679 \\
& CiaoSR~\cite{cao2023ciaosr} & 26.61/0.772 & 25.76/0.756 & 25.22/0.746 & 24.38/0.727 & 24.06/0.721 & 23.36/0.710 \\
& LMI~\cite{fu2024continuous} & 26.74/0.769 & 25.86/0.753 & 25.30/0.742 & 24.48/0.723 & 24.14/0.714 & 23.37/0.682 \\
& \bf{Ours} & \bf{26.74/0.777} & \bf{25.92/0.760} & \bf{25.34/0.749} & \bf{24.50/0.728} & \bf{24.15/0.724} & \bf{23.39/0.710} \\

\bottomrule
\end{tabular}
}
\end{table*}

\begin{figure*}[ht]
  \centering
  \includegraphics[width=\linewidth]{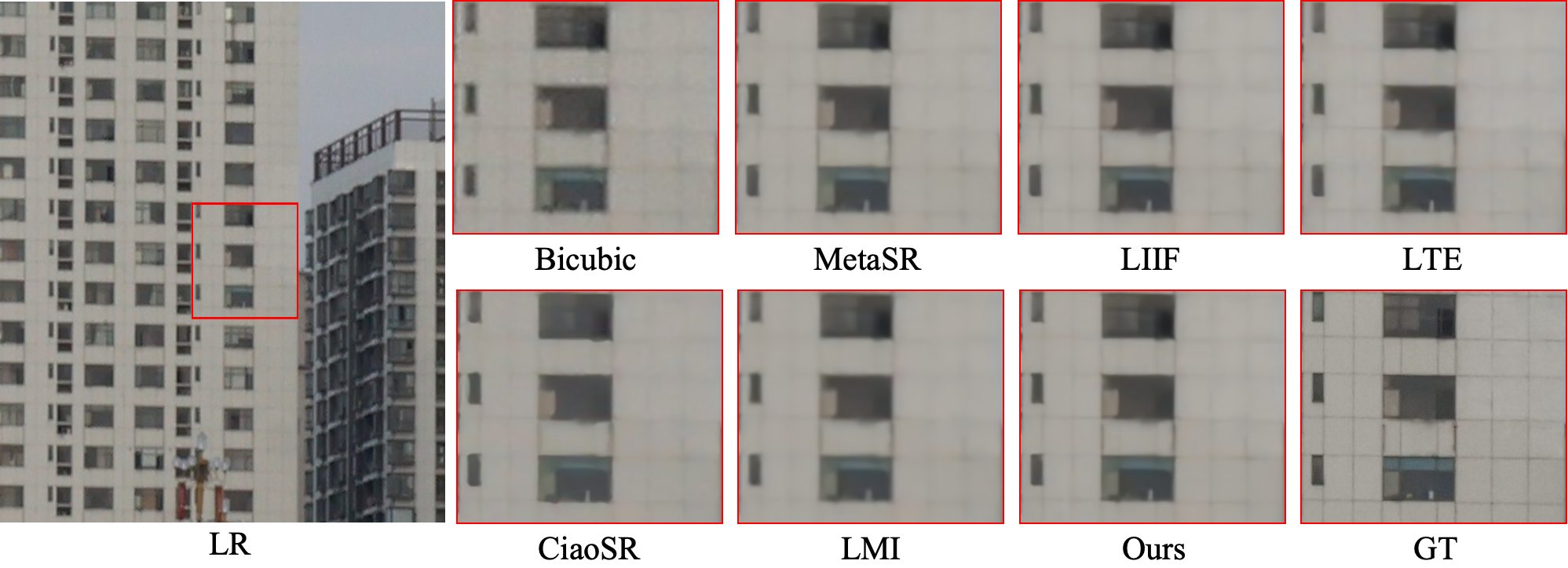}
\caption{Visual comparison with existing methods on the real COZ dataset $\times$ 3. Please zoom in for a better view.}
  \label{fig:COZ}
\end{figure*}

\subsection{Proposed $\text{S}^{3}$Mamba}

Further, to integrate global information and strengthen the scale-invariant perception capability of the feature space, we employ the SSSM as an efficient global feature extraction method to supplement global information. We also propose a novel scale-aware self-attention mechanism to further enhance the network's ability to perceive globally important features at different scales, as illustrated in Figure~\ref{fig:Method}. Specifically, for a low-resolution (LR) image, we first extract its features through a backbone, obtaining \( F_{LR} \). Additionally, we follow ~\cite{cao2023ciaosr} by using the Unfold operation to aggregate local features and obtain local information \( F^{\text{local}}_{LR} \). The SSSM is utilized to extract global features \( F^{\text{global}}_{LR} \). These are combined to form a new fused feature for subsequent representation learning, as shown in the following equations:
\begin{equation}
\begin{aligned}
    F^{local}_{LR},  F^{global}_{LR} &= U(F_{LR}) , SSSM(F_{LR}), \\
    F_{fusion} &= concat(F^{local}_{LR}, F^{global}_{LR}). \\    
\end{aligned}
\label{eq:9}
\end{equation}
where $U$ represents the unfold operation to capture local features. In addition, considering the inconsistency in feature distribution across different scales, we propose a scale-aware self-attention mechanism to enhance the network's focus on the feature representation at the current scale. This mechanism aims to learn a feature-independent global mapping pattern under various transformation modes. Specifically, we input \({coord}_{HR}\) and \({scale}\) into the SSSM to generate a global self-attention map \(\alpha_{{weight}}\). This attention map, guided by the current scale and coordinates, adaptively refines high-resolution feature \(F_{HR}\), ultimately yielding \({RGB}_{HR}\). The process is illustrated by the following equations:
\begin{equation}
\begin{aligned}
    \alpha_{{weight}} &= {SSSM}({coord}_{{HR}}, {scale}), \\  
    F'_{{HR}} &=  {SSSM}(\alpha_{{weight}} \cdot F_{{HR}}), \\
    {RGB}_{{HR}} &= {SSSM}(F'_{{HR}}).
\end{aligned}
\label{eq:10}
\end{equation}
Finally, based on these, we build a simple yet efficient arbitrary-scale super-resolution architecture called $\text{S}^{3}$Mamba, as illustrated in Figure~\ref{fig:Method} (a).

\begin{table*}[ht]
\caption{Quantitative comparison with state-of-the-art methods for arbitrary-scale SR on the \underline{\textbf{synthetic DIV2K validation set (PSNR (dB))}}. \textbf{Bold} and \underline{Bold} indicate the best performance and second-best performance, respectively. Out-of-scale means the models were not trained with these large scales.
}
\label{tab:comp_DIV2K}
\centering
\resizebox{0.95\textwidth}{!}{
\begin{tabular}{l||l||ccc||ccccc}
\toprule
& & \multicolumn{3}{c||}{$\qquad$In-scale$\qquad$} & \multicolumn{5}{c}{Out-of-scale} \\
\cmidrule{3-10}
\multirow{-2}{*}{Backbones}&  \multirow{-2}{*}{Methods} &~~~~~$\times2$~~~~~ & ~~~~~$\times3$~~~~~ & ~~~~~$\times4$~~~~~ & ~~~$\times6$~~~ & ~~~$\times12$~~~ & ~~~$\times18$~~~ & ~~~$\times24$~~~ & ~~~$\times30$~~~ 
\\
\midrule
\multirow{1}{*}{} &
{Bicubic} & 31.01 & 28.22 & 26.66 & 24.82  & 22.27 & 21.00 & 20.19 & 19.59 \\
\hline
\multirow{9}{*}{EDSR~\cite{lim2017edsr}} &
EDSR-baseline~\cite{lim2017edsr} & 34.55 & 30.90 & 28.94 & - & - & - & - & - \\
& MetaSR~\cite{hu2019metasr} & 34.64 & 30.93 & 28.92 & 26.61 & 23.55 & 22.03 & 21.06 & 20.37 \\
& LIIF~\cite{chen2021liif} & {34.67} & {30.96} & {29.00} & {26.75} & {23.71} & {22.17} & {21.18} & {20.48} \\
& ITSRN~\cite{yang202itsrn} &
34.71 & 30.95 & 29.03 & 26.77 & 23.71 & 22.17 & 21.18 & 20.49 \\
& LMI~\cite{fu2024continuous} & 34.59 & 30.90 & 28.94 & 26.69 & 23.68 & 22.18 & 21.23 & 20.55 \\
& LTE~\cite{lee2022lte} & {34.72} & {31.02} & {29.04} & {26.81} & {23.78} & {22.23} & {21.24} & {20.53} \\
& CLIT~\cite{chen2023cascaded} & 34.81 & 31.12 & 29.15 & 26.92 & 23.83 & 22.29 & 21.26 & 20.53\\
& SRNO~\cite{wei2023super} & 34.85 & 31.11 & 29.16 & 26.90 & 23.84 & 22.29 & 21.27 & 20.56\\
& CiaoSR~\cite{cao2023ciaosr} & \underline{34.91} & \bf{31.15} & \underline{29.23} & \underline{26.95} & \underline{23.88} & \underline{22.32} & \bf{21.32} & \underline{20.59}  \\

& \bf{Ours} & \bf{34.93} & \underline{31.13} & \bf{29.24} & \bf{26.97} & \bf{23.89} & \bf{22.32} & \underline{21.30} & \bf{20.59} \\

\midrule
\multirow{9}{*}{RDN~\cite{zhang2018rdn}} &
RDN-baseline~\cite{zhang2018rdn} & 34.94 & 31.22 & 29.19 & - & - & - & - & - \\
& MetaSR~\cite{hu2019metasr} & {35.00} & {31.27} & 29.25 & 26.88 & 23.73 & 22.18 & 21.17 & 20.47 \\
& LIIF~\cite{chen2021liif} & 34.99 & 31.26 & {29.27} & {26.99} & {23.89} & {22.34} & {21.31} & {20.59} \\
& ITSRN~\cite{yang202itsrn} &
35.09 & 31.36 & 29.38 & 27.06 & 23.93 & 22.36 & 21.32 & 20.61 \\
& LMI~\cite{fu2024continuous} & 34.74 & 31.03 & 29.07 & 26.81 & 23.79 & 22.29 & 21.31 & 20.63 \\
& LTE~\cite{lee2022lte} & {35.04} & {31.32} & {29.33} & {27.04} & {23.95} & {22.40} & {21.36} & {20.64} \\
& CLIT~\cite{chen2023cascaded} & 35.10 & 31.39 & 29.39 & 27.12 & 24.01 & 22.45 & 21.38 & 20.64\\
& SRNO~\cite{wei2023super} & \underline{35.16} & {31.42} & 29.42 & 27.12 & 24.03 & 22.46 & 21.41 & 20.68\\
& CiaoSR~\cite{cao2023ciaosr} & 35.15 & \bf{31.42} & \underline{29.45} & \underline{27.16} & \underline{24.06} & \underline{22.48} & \underline{21.43} & \bf{20.70}  \\
& \bf{Ours} & \bf{35.17} & \underline{31.40} & \bf{29.47} & \bf{27.17} & \bf{24.07} & \bf{22.50} & \bf{21.43} & \underline{20.68} \\
\bottomrule
\end{tabular}
}
\end{table*}

\begin{figure*}[ht]
  \centering
  \includegraphics[width=\linewidth]{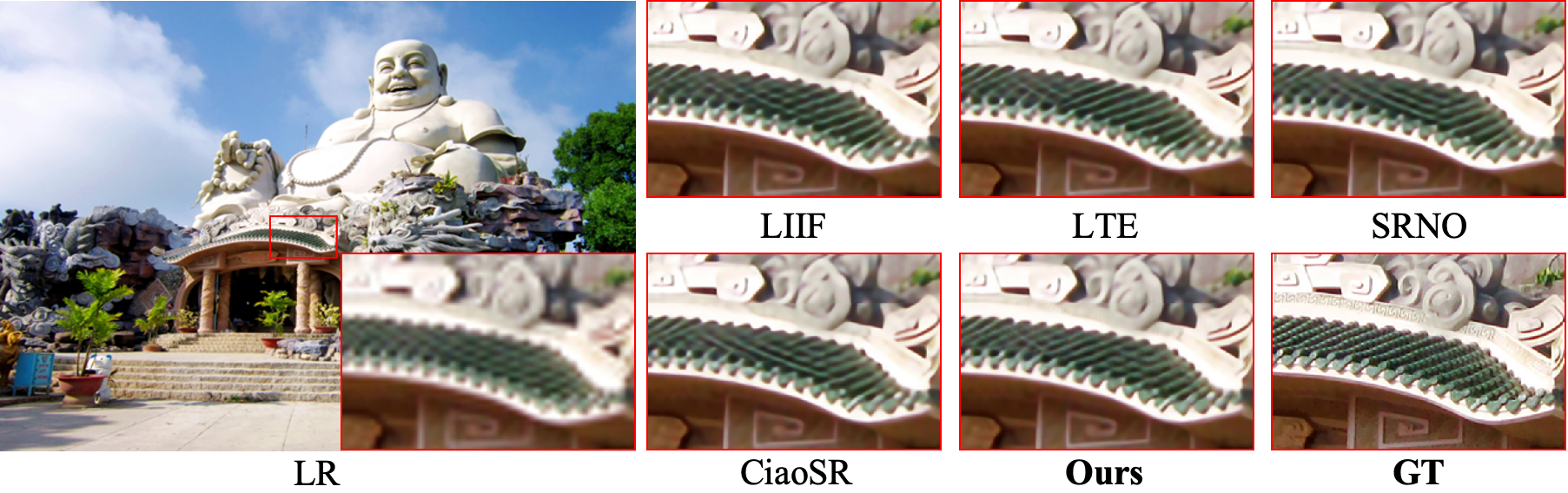}
\caption{Visual comparison with existing methods on the DIV2K dataset $\times$ 4. Please zoom in for a better view.}
  \label{fig:DIV2K}
\end{figure*}

\section{Experiment and Analysis}
\label{sec:exp}
\subsection{Experiments Setting}


{\bf Datasets.} For evaluating the real-world arbitrary-scale SR task, we follow the training and test set from COZ~\cite{fu2024continuous}, which consists of 153 images at 2K resolution for training and 37 images at 2K resolution for testing. Following previous works~\cite{chen2021liif,LTE}, we also use the commonly employed synthetic DIV2K~\cite{Agustsson_2017_CVPR_Workshops} dataset as the training set, which consists of 800 high-resolution (HR) images in 2K resolution for training by bicubic degradation model. For testing on DIV2K, we evaluate the performance of different models on the DIV2K validation set with 800 high-resolution (HR) images in 2K resolution.\\
{\bf Evaluation metrics.} Following previous work~\cite{chen2021liif,lee2022lte}, we use Peak Signal-to-Noise Ratio (PSNR) and Structural Similarity Index (SSIM)~\cite{SSIM} to evaluate the quality of the generated HR images. Note that the PSNR value is calculated on the RGB channels for the DIV2K validation set and on the Y channel (\ie, luminance) of the transformed YCbCr space for the other benchmark test sets.\\
{\bf Implementation details.} Following previous works~\cite{chen2021liif,lee2022lte}, we adopt the same way to generate paired images for training arbitrary-scale super-resolution models. Specifically, initially, we crop image patches of size 96\(s\) $\times$ 96\(s\) as ground truth (GT), where \(s\) is a scaling factor sampled from a uniform distribution U(1, 4). Then, we use bicubic downsampling to generate corresponding low-resolution (LR) images. We employ existing SR models, such as EDSR~\cite{lim2017edsr} and RDN~\cite{zhang2018rdn}, as backbones to evaluate various arbitrary-scale upsampling methods. Adam~\cite{Adam} is used as the optimizer, with the initial learning rate set to 1e-4 and decaying by a factor of 0.5 every 200 epochs. During training, our method follows~\cite{chen2021liif,Meta-SR,lee2022lte}, training for a total of 1000 epochs under L1 loss. The total batch size is set to 32, utilizing a total of 8 V100 GPUs. For real-world arbitrary-scale SR datasets~\cite{fu2024continuous}, we follow the training and testing setting of COZ~\cite{fu2024continuous} to ensure fair evaluation, while the total batch size is set to 128.\\
{\bf Compared methods.} To demonstrate the superiority of our model, we conduct a performance comparison against eight state-of-the-art (SOTA) and popular models: MetaSR~\cite{hu2019metasr}, LIIF~\cite{chen2021liif}, LTE~\cite{lee2022lte}, LINF~\cite{yao2023local}, SRNO~\cite{wei2023super}, LIT~\cite{chen2023cascaded}, CiaoSR~\cite{cao2023ciaosr}, and LMI~\cite{fu2024continuous} under two popular backbones EDSR~\cite{lim2017edsr} and RDN~\cite{zhang2018rdn}.

\subsection{Quantitative and Qualitative results}
To demonstrate the superiority of our method, we conduct a comparative experiment on the COZ dataset and DIV2K datasets to assess its performance against existing methods, as shown in Tables~\ref{tab:comp_COZ} and ~\ref{tab:comp_DIV2K}.\\
{\bf Results on the COZ dataset.} As shown in Table~\ref{tab:comp_COZ}, we can observe that compared to other approaches, our method achieves significant improvements on real-world COZ datasets, particularly with a notable enhancement in the SSIM metric. For instance, in the scale $\times$3.5 on the RDN baseline, our method surpasses existing SOTA methods by 0.06db in PSNR and 0.004 in SSIM. This demonstrates that our method is capable of better reconstructing continuous image representations, thereby enhancing image detail performance. Additionally, we conduct a visual comparison of the COZ dataset, as shown in Figure~\ref{fig:COZ}. It is evident that compared with existing methods, our method more effectively removes degradation artifacts in real-world scenarios, reconstructing the details and textures of super-resolution images closer to GT. This demonstrates the superior performance of the proposed method in real-world applications.\\
{\bf Results on the DIV2K dataset.} As shown in Table~\ref{tab:comp_DIV2K}, compared to existing methods, our method achieves the best performance in most scenarios. For instance, our method surpasses all existing methods in scenarios like $\times$2, $\times$4, \textit{etc.}, achieving the best performance. Additionally, our method also achieves the best performance in most out-of-scale scenarios. Although the performance of CiaoSR is comparable to our method, our computational complexity is only half of its, as shown in the supplementary materials. Furthermore, we conduct a visual comparison of the DIV2K, as shown in Figure~\ref{fig:DIV2K}. It can be observed that the texture details reconstructed by our method are closer to the ground truth (GT), whereas other methods, such as the existing state-of-the-art method CiaoSR, tend to produce artifacts. This demonstrates the superior visual performance of our approach.

\begin{table}[t]
\centering

\resizebox{1.0\linewidth}{!}{
\begin{tabular}{ccc|c|c}
\toprule 
MLP & SSM  & Our SSSM & PSNR on \texttimes2 & PSNR on \texttimes4 \\ \midrule 
\ding{51} & \ding{55} & \ding{55} & 34.78 & 29.09 \\ 
\ding{55} & \ding{51} & \ding{55} & 34.85 & 29.17 \\  
\ding{55} & \ding{55} & \ding{51} & \bf{34.91} & \bf{29.24} \\ \bottomrule
\end{tabular}}
\caption{Performance comparison of the different base models.}
 \vspace{-5mm}
\label{tab:ab_SSSM}
\end{table}

\subsection{Ablation Study}
In this section, we conduct ablation studies to evaluate the effectiveness of the core ideas of our method. We focus on two components: (a) the Scalable State Space Model and (b) key elements in $\text{S}^{3}$mamba, global feature extraction (GFE) and scale-aware self-attention (SFAtt). We use the EDSR baseline to validate their effectiveness on the DIV2K dataset.\\
{\bf Effectiveness of SSSM.}  We propose the Scalable State Space Model (SSSM) to facilitate global continuous modeling. To evaluate it, we conduct comparisons by replacing our proposed SSSM with the MLP and traditional SSM module, and the results are shown in Table~\ref{tab:ab_SSSM}. We can observe that the model incorporating traditional SSM with the SSM structure significantly outperforms the MLP-based model, confirming the efficacy of SSM in capturing global information. Furthermore, after equipping it with scale and position awareness, our proposed Scalable State Space Model surpasses the traditional SSM block and MLP, demonstrating superior performance.\\
\noindent {\bf Effectiveness of GFE and SFAtt.} The $\text{S}^{3}$mamba includes several core modules: global feature extraction (GFE) and scale-aware self-attention (SFAtt). To verify the effectiveness of these modules, we remove them for performance comparison, as shown in Table~\ref{tab:ab_GFE_Att}. Specifically, after integrating SFAtt, the network's performance improved by 0.07 dB in PSNR for \texttimes2, indicating that this attention mechanism facilitates the perception of different scales. Additionally, the inclusion of GFE further enhanced the network's performance by 0.06 dB in PSNR for \texttimes2, demonstrating that global information is crucial for learning a continuous representation space. Finally, by combining these two core modules, our method achieves the best performance.

\begin{table}[t]
\centering

\resizebox{1.0\linewidth}{!}{
\begin{tabular}{cc|c|c|c}
\toprule 
GFE & SFAtt & PSNR on \texttimes2 & PSNR on \texttimes3  & PSNR on \texttimes4 \\ \midrule
\ding{55} & \ding{55} & 34.71 & 30.98 & 29.06 \\ 
\ding{55} & \ding{51} & 34.78 & 31.03 & 29.12 \\ 
\ding{51} & \ding{55} & 34.85 & 31.09 & 29.19 \\
\ding{51} & \ding{51} & \bf{34.91}  & \bf{31.13} & \bf{29.24} \\ \bottomrule
\end{tabular}}
\caption{Performance comparison of the proposed core modules GFE and SFAtt on the DIV2K dataset.}
\label{tab:ab_GFE_Att}
 \vspace{-4mm}

\end{table}
For a comprehensive understanding of our proposed $\text{S}^3$Mamba framework, we provide additional discussions, comparisons, detailed analyses, and extensive visual examples in the Appendix, highlighting its superiority.

\section{Conclusion}
In this paper, we propose a novel Scalable State Space Model (SSSM) that modulates the state transition and sampling matrices during the discretization process, achieving scalable and continuous representation modeling with linear computational complexity. Additionally, we develop a novel scale-aware self-attention mechanism to further enhance the network's ability to perceive globally significant features across various scales. The $\text{S}^{3}$Mamba is designed for constructing scalable continuous representation spaces, enabling the reconstruction of arbitrary-scale high-resolution images with rich detail. Extensive experiments on both synthetic and real-world benchmarks demonstrate that our method not only achieves state-of-the-art results but also exhibits remarkable generalization capabilities, paving the new way for arbitrary-scale super-resolution.
\clearpage
{
    \small
    \bibliographystyle{ieeenat_fullname}
    \bibliography{main}
}


\end{document}